\documentclass{ecai}
\usepackage{times}
\usepackage{graphicx}
\usepackage{latexsym}
\usepackage{url}
\usepackage{hyperref}
\usepackage{amsmath, amssymb}
\usepackage[ruled]{algorithm2e}
\usepackage[T3,OT2,T1]{fontenc} 
\usepackage[noenc]{tipa}
\usepackage{dblfloatfix}
\usepackage{tikz}
\usepackage{multirow} 
\usepackage{enumitem} 
\newcommand{\subscript}[2]{$#1 _ #2$} 

\ecaisubmission

\begin{document}

\title{LionForests: Local Interpretation of Random Forests}

\author{Ioannis Mollas, Nick Bassiliades, Ioannis Vlahavas, Grigorios Tsoumakas\institute{Department of Informatics, Aristotle University of Thessaloniki, 54124 Thessaloniki, Greece, email: \{iamollas,nbassili,vlahavas,greg\}@csd.auth.gr} }

\maketitle
\bibliographystyle{ecai}

\begin{abstract}
Towards a future where ML systems will integrate into every aspect of people's lives, researching methods to interpret such systems is necessary, instead of focusing exclusively on enhancing their performance. Enriching the trust between these systems and people will accelerate this integration process. Many medical and retail banking/finance applications use state-of-the-art ML techniques to predict certain aspects of new instances. Thus, explainability is a key requirement for human-centred AI approaches. Tree ensembles, like random forests, are widely acceptable solutions on these tasks, while at the same time they are avoided due to their black-box uninterpretable nature, creating an unreasonable paradox. In this paper, we provide a methodology for shedding light on the predictions of the misjudged family of tree ensemble algorithms. Using classic unsupervised learning techniques and an enhanced similarity metric, to wander among transparent trees inside a forest \emph{following breadcrumbs}, the interpretable essence of tree ensembles arises. An interpretation provided by these systems using our approach, which we call ``LionForests'', can be a simple, comprehensive rule.
\end{abstract}

\section{INTRODUCTION}
Machine learning (ML) models are becoming pervasive in our society and everyday life. Such models may contain errors, or may be subject to manipulation from an adversary. In addition, they may be mirroring the biases that exist in the data from which they were induced. For example, Apple's new credit card is being recently investigated over claims it gives women lower credit~\cite{womanCard}, IBM Watson Health was accused of suggesting unsafe treatments for patients~\cite{watsonAI} and state-of-the-art object detection model YOLOv2 is easily tricked by specially designed patches~\cite{pattern1,pattern2}. Being able to understand how an ML model operates and why it predicts a particular outcome is important for engineering safe and unbiased intelligent systems. 

Unfortunately, many families of highly accurate (and thus popular) models, such as deep neural networks and tree ensembles, are opaque: humans cannot understand the inner workings of such models and/or the reasons underpinning their predictions. This has recently motivated the development of a large body of research on {\em interpretable ML} (IML), concerned with the interpretation of black box models~\cite{adadi,dovsilovic,duliu,rivardo,tameru,mythosOfInter,samek}.

Methods for interpreting ML models are categorised, among other dimensions~\cite{rivardo}, into {\em global} ones that uncover the whole logic and structure of a model and {\em local} ones that aim to interpret a single prediction, such as ``Why has this patient to be immediately hospitalized?''. This work focuses on the latter category. Besides their utility in uncovering errors and biases, local interpretation methods are in certain domains a prerequisite due to legal frameworks, such as the General Data Protection Regulation (GDPR)~\cite{gdpr} of the EU and the Equal Credit Opportunity Act of the US\footnote{ECOA 15 U.S. Code \S1691 et seq.: \url{https://www.law.cornell.edu/uscode/text/15/1691}}.

Another important dimension, which IML methods can be categorised, concerns the type of ML model that they are interpreting~\cite{adadi}. {\em Model-agnostic} methods~\cite{shap, lime, anchors} can be applied to any type of model, while {\em model-specific} methods~\cite{lrp, shapTrees, lionets, moore} are engineered for a specific type of model. Methods of the former category have wider applicability, but they just approximately explain the models they are applied to~\cite{adadi}. This work focuses on the latter category of methods, proposing a technique specified for tree ensembles~\cite{randomForests, xgboost}, which are very effective in several applications involving tabular and time-series data~\cite{rfrwa}. 

Past work on model-specific interpretation techniques, about tree ensembles, is limited~\cite{moore,iforest}. iForest~\cite{iforest}, a global and local interpretation system of random forests (RF), provides insights for a decision through a visualisation tool. Notwithstanding, such visual explanation tool is very complex for non-expert users, while at the same time requires user interaction in order to construct the local interpretations. Another instance-level interpretation technique for RF~\cite{moore}, produces an interpretation in the form of a list of features with their ranges, accompanied by an influence metric. If the list of features is extensive, and the ranges are very narrow, the interpretation can be considered as unreliable and untrustworthy, because small changes in the features will render the interpretation useless. Finally, both methods do not handle categorical data appropriately.

To address the above problems, we introduce a local-based model-specific approach for interpreting an individual prediction of an RF through a single rule, in natural language form. The ultimate goal is mainly to reduce the number of features and secondly to broaden the feature-ranges producing more robust, indisputable and intuitive interpretations. Additionally, the categorical features are handled properly, providing intelligible information about them throughout the rules. The constructed rule will be presented as the interpretation, if its length is acceptable to be comprehensible. Otherwise, additional processing will be held to form an acceptable interpretation. We call this technique ``LionForests'' (Local Interpretation Of raNdom FORESTS) and we use its path and feature selection ability, based on unsupervised techniques like association rules~\cite{associationrules} and $k$-medoids clustering~\cite{kmedoidsori} to process the interpretations in order to make them more comprehensible.

The rest of this work is structured the following way. First, Section 2 introduces the related work, presenting approaches of interpretation techniques applicable to RF models. In Section 3, we present the methodology of our approach to the interpretation of RF models. In section 4, we perform the quantitative and qualitative analysis. Finally, we conclude and present possible future directions of our strategy in Section 5.

\section{RELATED WORK}

A number of works concerning the tree ensembles interpretation problem are either model-agnostic or model-specific solutions, with a global or local scope, as presented with a similar taxonomy in a recent survey~\cite{haddouchi2019survey}. 

A family of model-agnostic interpretation techniques about black-box models, including tree ensembles, concerns the efficient calculation of feature importance. These are variations of feature permutation methods~\cite{permFI}, partial dependence plots~\cite{pdp} and individual conditional expectation~\cite{iceplots}, which are global-based. SHAP~\cite{shap} is an alternative method to compute feature importance for both global and local aspects of any black-box model.

Specifically, on tree ensembles, the most common techniques include the processes of extracting, measuring, pruning and selecting rules from the trees to compute the feature importance~\cite{inTrees, otherFIontrees}. A highly studied by many researchers~\cite{inTrees,domingos,satoshiTrees,zhouTrees} technique attempts to globally interpret tree ensembles using single-tree approximations. But this method, as its name implies, approximates the performance of the model it seeks to explain. Thus, this approach is extremely problematic and criticised, because it is not feasible to summarise a complex model like tree ensembles to a single tree~\cite{donotusesingletreesappx}. An additional approach on interpreting tree ensembles focuses on clustering the trees of an ensemble using a tree dissimilarity metric~\cite{chipman1998making}. 

\begin{figure}
\centerline{\includegraphics[height=1.1in]{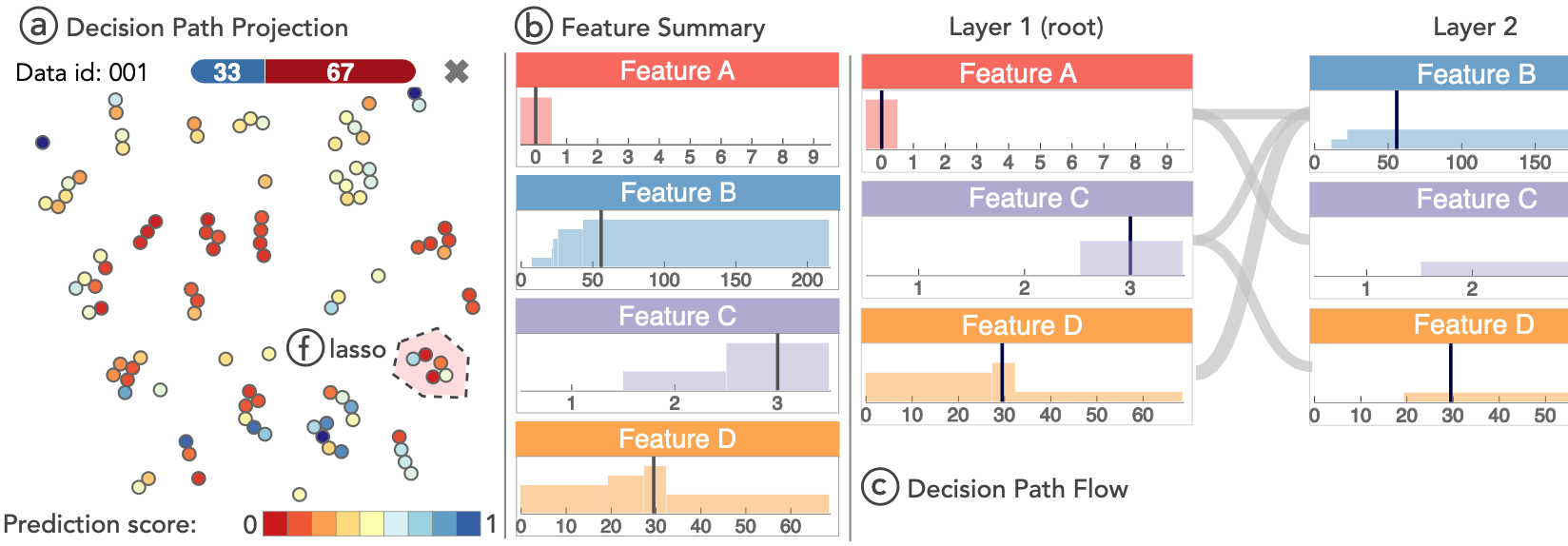}}
\caption{A sample example from the visualisation tool of iForest adapted from~\cite{iforest}} \label{fig:iForest}
\end{figure}

iForest~\cite{iforest}, a model-specific global and local approach, utilises a path distance metric to project an instance's paths to a two-dimensional space via t-Distributed Stochastic Neighbour Embedding (t-SNE)~\cite{tsne}. The path distance metric they propose considers distant two paths in two cases: a) if a feature exists in only one out of the two paths the distance between those two paths is increasing, b) if a feature exists in both paths, the distance is increasing according to the non-common ranges of the feature on the paths divided by half. The total distance, which is the aggregation of those cases for all the features, is finally divided by the total number of features appearing at least in one out of the two paths. Except the projection of the paths (Figure~\ref{fig:iForest}a), they provide feature summary (Figure~\ref{fig:iForest}b) and decision path flow (Figure~\ref{fig:iForest}c), which is a paths overview. In feature summary, a stacked area plot visualises every path's range for a specific feature, while decision path flow plot visualises the paths themselves. However, they do not provide this information automatically. The user has to draw a lasso (Figure~\ref{fig:iForest}f) around some points-paths in the paths projection plot in order to get the feature summary and paths overview. But requiring the user to select the appropriate paths is critical, simply because the user can easily choose wrong paths, a small set of paths, or even paths entirely different to the paths being responsible for his prediction. That may lead to incorrect feature summary and paths overview, thus to a faulty interpretation.

\begin{figure}
\centerline{\includegraphics[height=1in]{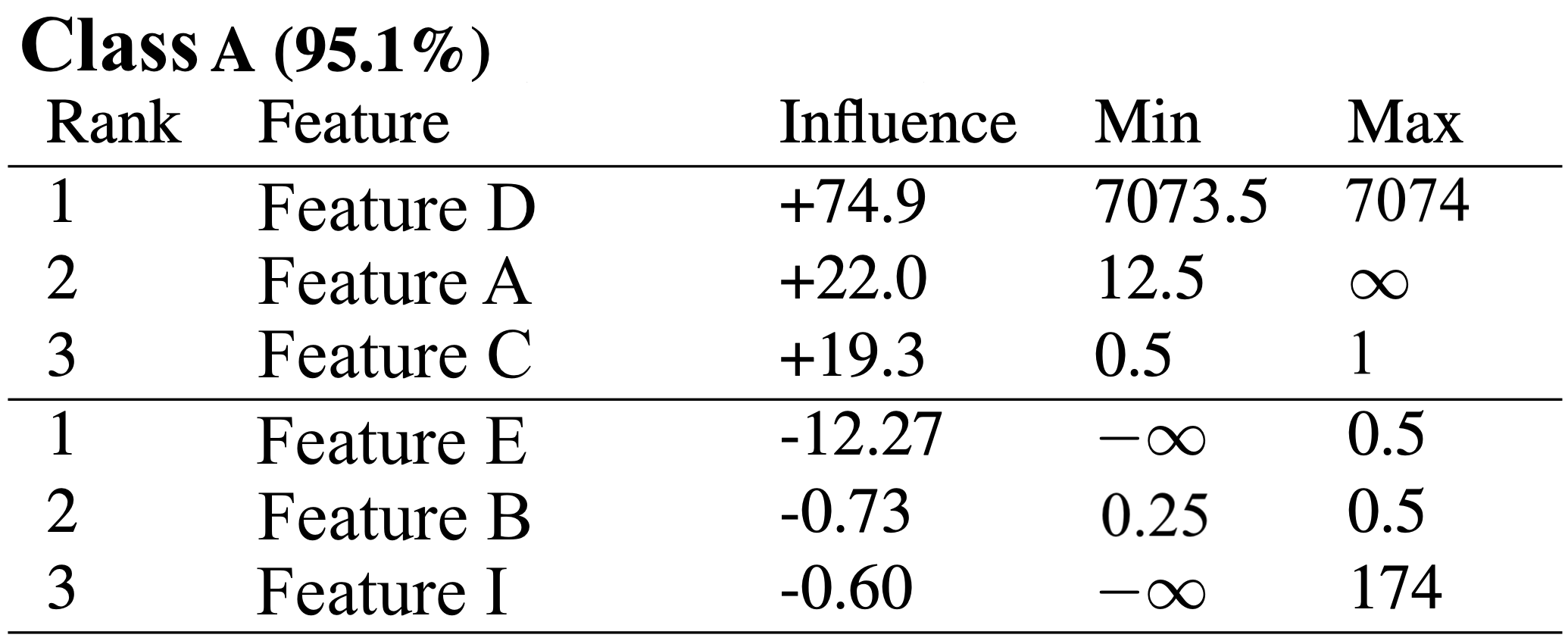}}
\caption{A sample explanation from~\cite{moore}} \label{fig:moore}
\end{figure}

Lastly, one technique~\cite{moore} interprets tree ensembles in instance-level (local-based technique) providing as interpretation a set of features with their ranges, ranked based on their contribution (see Figure~\ref{fig:moore}). Thus, the interpretation process consists of two parts. Firstly, they calculate the influence of a feature by monitoring the changes of the activated nodes for a specific instance's prediction. This influence later will be used for the ranking process. The second step is to find the narrowest range across all trees for every feature. However, they do not attempt to expand these ranges, while they also claim that their influence metric assigns zero influence to some features, and by extension removing them, they could offer more compact explanations. In spite of this, they do not know, by keeping only features with a non-zero influence, that these features will at least be present in half plus one paths to preserve the same prediction of an instance. Finally, they do not manage categorical features properly.

\section{OUR APPROACH} 
Our objective is to provide local interpretation of RF binary classifiers. In RF, a set of techniques like data and feature sampling, is used in order to train a collection of $T$ weak trees. Then, these trained trees vote for an instance's prediction: \begin{equation} \label{eq:1} h(x_i) = \dfrac{1}{|T|} \sum_{t \in T}^{} h_t(x_i) \end{equation} where $h_t(x_i)$ is the vote cast from the tree $t \in T$ for the instance $x_i \in X$, representing the probability $P(C=c_j|X=x_i)$ of $x_i$ to be assigned to class $c_j \in C = \{0,1\}$, thus \[ h_t(x_i) =
  \begin{cases}
   1       & \quad \text{if } P(C=1|X=x_i) \geq 0.5 \\
   0       & \quad \text{if } P(C=0|X=x_i) \geq 0.5.
  \end{cases}
\]

\begin{figure}
\centerline{\includegraphics[height=1.7in]{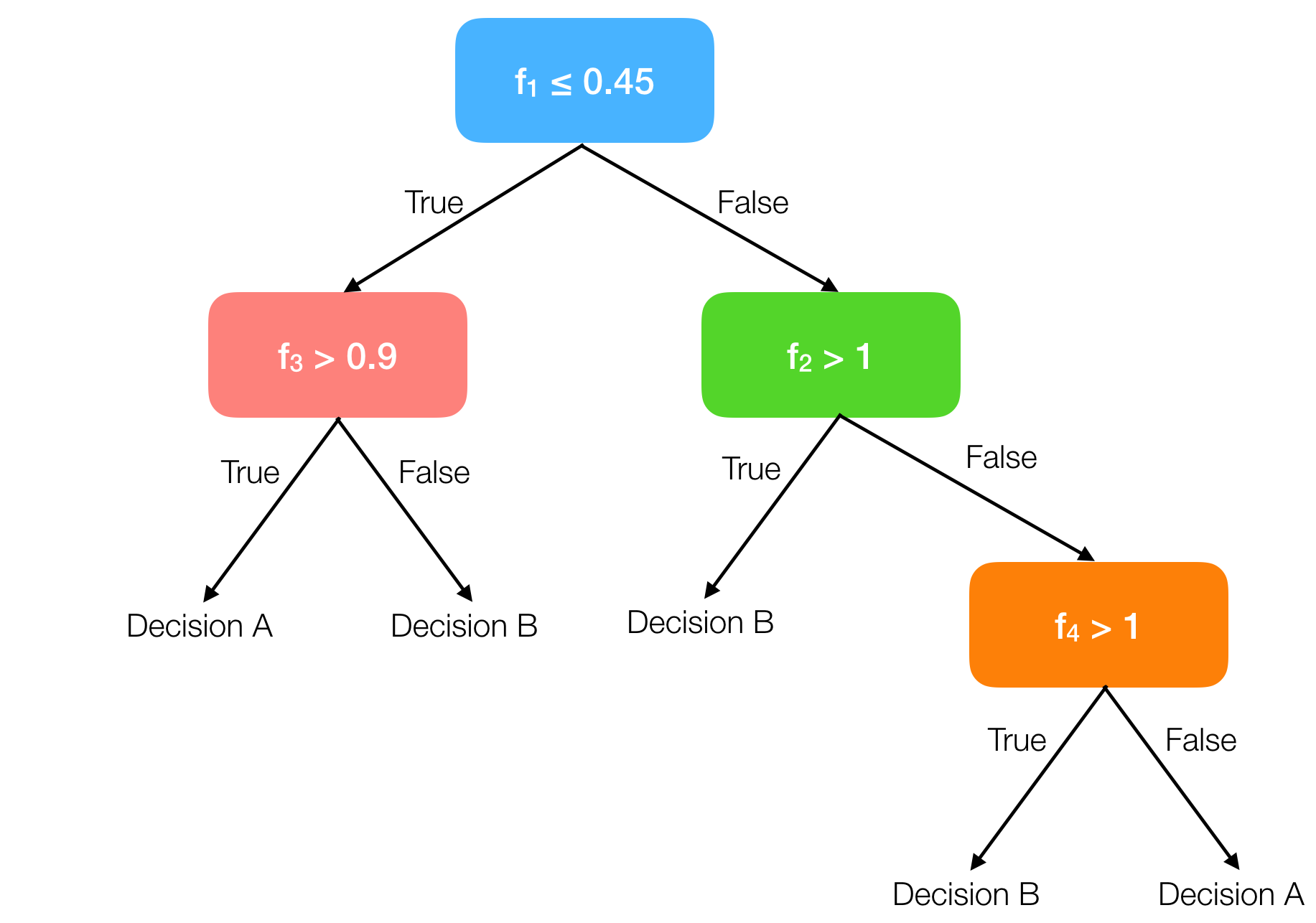}}
\caption{A decision tree binary classifier with 4 features} \label{exampledt}
\end{figure}

Each decision tree $t \in T$ is a directed graph, and by deconstructing its structure, we are able to derive a set $P_t$ of paths from the root to the leaves. Therefore, every instance can be classified with one of these paths. A path $p \in P_t$ is a conjunction of conditions, and the conditions are features and values with relations $\leq$ and $>$. For example, a path from the tree on Figure~\ref{exampledt} could be the following: `if $f_1 \leq 0.45$ and $f_3 > 0.9$ then Decision A'. Thus, each path $p$ is expressed as a set:\begin{equation}p=\{f_i \boxtimes v_j | f_i \in F, v_j \in \Re, \boxtimes \in \{\leq, >\}\}\end{equation}

We are presenting LionForests, a framework for interpreting RF models in the instance level. LionForests is a pipeline of actions: a) feature-ranges extraction, reduction through b$_1$) association rules, b$_2$) clustering and b$_3$) random selection, c) categorical feature handling, and, finally, d) interpretation composition.

\subsection{Feature-Ranges Extraction}

\begin{figure}[t]
\centerline{\includegraphics[height=2in]{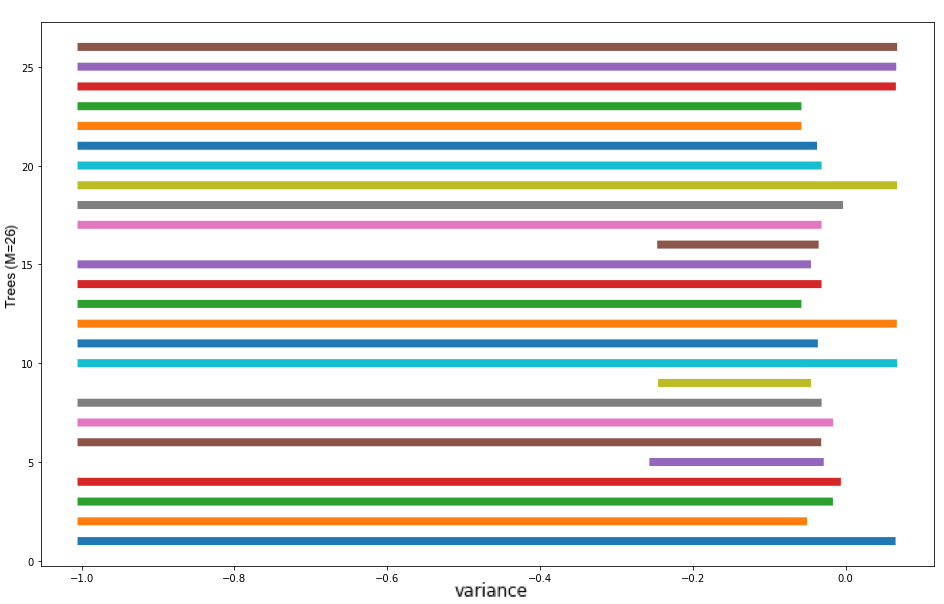}}
\caption{Bars plot of Value ranges of the `variance' feature from the Banknote Dataset for the $M=26$ trees of an RF of $K=42$ trees predicting the majority class} \label{BarsPlot}
\end{figure}

\begin{figure}[t]
\centerline{\includegraphics[height=2in]{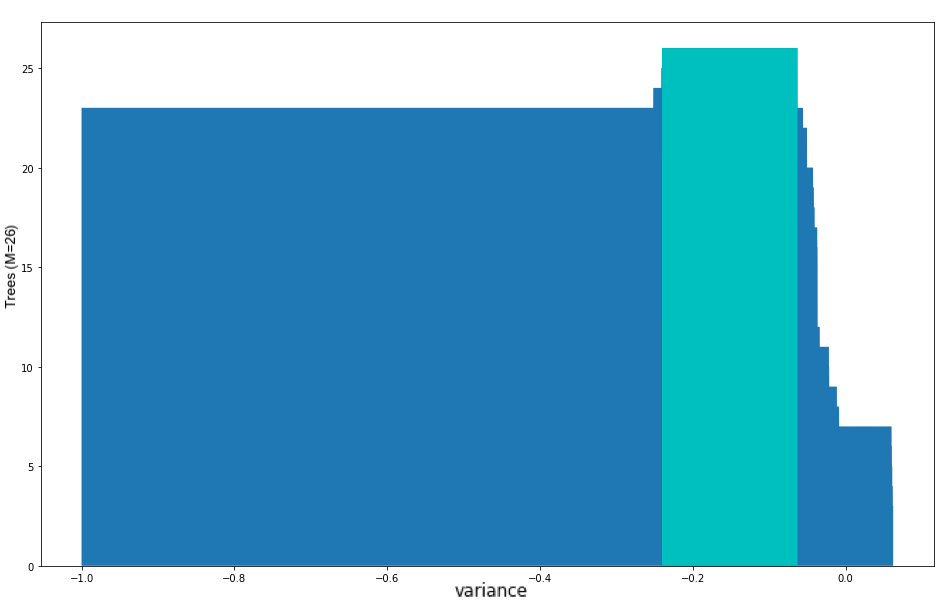}}
\caption{Stacked area plot of value ranges of the `variance' feature from the Banknote Dataset for the $M=26$ trees of an RF of $K=42$ trees predicting the majority class} \label{stackedAreaPlot}
\end{figure}

Our approach delivers local explanations for RF. Assume we have an RF of $N$ trees and an instance $x$, which is classified by the forest as $c_j \in C =\{0,1\}$. We focus on the $K \geq \frac{N}{2}$ trees of the forest that classify $x$ as $c_j$. For each feature, we compute from each of the $K$ trees that it appears in, its values range as imposed from the conditions involving this feature in the path from the root to the leaf. Figure~\ref{BarsPlot} shows an example of these ranges for feature `variance' with range $-1...0.1$ from the Banknote dataset~\cite{ucidata} for a particular RF of $50$ trees and a particular instance whose value for the skew feature is $-0.179$. Figure~\ref{stackedAreaPlot} shows the corresponding stacked area plot. The highlighted (cyan/light grey) area represents the intersection of these ranges, which they will always contain the instance's value for the specific feature. Moreover, no matter how much the feature value is going to change, as long as it stays within this intersection range, the instance will not follow a different decision path in these trees.

In order to give a brief example, we have the following three paths:

\begin{enumerate}[label=\subscript{p}{{\arabic*}}]
    \item if $f_1 \leq 0.6$ and $...$ then Class\_A
    \item if $f_1 \leq 0.6$ and $f_1 > 0.469$ and $...$ then Class\_A
    \item if $f_1 > 0$ and $f_1 \leq 1$ and $...$ then Class\_A
\end{enumerate}

\begin{figure}[h]
\setlength\abovecaptionskip{0pt}
\setlength\belowcaptionskip{-0.3\baselineskip}
\centerline{\includegraphics[height=1in]{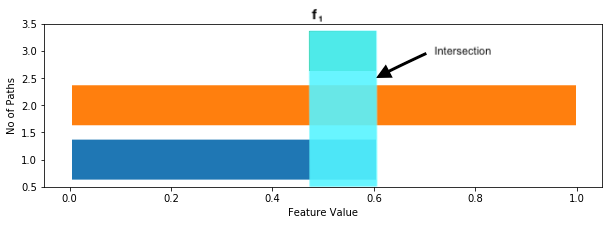}}
\caption{Bar plot of the simple example} \label{exampleBarPlot}
\end{figure}

The intersection of these three paths is the cyan/light grey area in Figure~\ref{exampleBarPlot}, which we can infer that the $f_1$ feature's range can be: $0.47 \leq f_1 \leq 0.6$. This intersection range will always contain the instance's value for the specific feature. Moreover, no matter how much the feature value is going to change, as long as it stays within this intersection range, the decision paths are not going to change. For example, if the value of the instance for the feature $f_1$ was $0.5$, if the value will change to $0.52$, each tree will take its decision through the same path. Summarising the aforementioned, an interpretation can have this shape:

\begin{quote}
\centering
    \smaller{`if $0.47 \leq f_1 \leq 0.6$ and $\dots$ then Class\_$A$'}
\end{quote}

But with as many paths as possible to vote to class $A$, we face the following two issues:

\begin{enumerate}
    \item A lot of paths can lead to an explanation with many features, by extension to an unintelligible understanding and a frustrated user.
    \item A lot of paths will lead to a small, strict and very specific feature range. For example, $f_1$ instance's value was $0.5$ and the intersection range of all paths for this feature occurs to be $0.47$$\leq f_1$$\leq 0.6$, while the feature range is $[-1,1]$. A narrow range, like the aforementioned, would result in a negative impression of the model, which will be considered unstable and unreliable. Then, a broader range will be less refutable.
\end{enumerate}

Consequently, we formulate the optimisation problem (Eq.~\ref{eq:optimazation}) to minimise the number of features that satisfy the paths of a subset of the original trees, thereby retaining the same classification result with the original set of trees and making the size of the total number of trees equal to or greater than the quorum, in order to ensure the consistency of the results of the original RF model.

\begin{equation}
\label{eq:optimazation}
\begin{aligned}
& \underset{F' \subseteq F}{\text{minimise}}
& & |F'|\\
& \text{subject to}
& & p=\{f_i \boxtimes v_j | f_i \in F'\},  p \in P_t \forall t \in T', \\
& \text{     }
& & \lfloor \frac{1}{|T|} \sum_{t \in T'}^{} h_t(x_i) + \frac{1}{2} \rfloor = \lfloor\frac{1}{|T|} \sum_{t \in T}^{} h_t(x_i) + \frac{1}{2}\rfloor,\\
& \text{     }
& & |T'| \geq quorum
\end{aligned}
\end{equation}

To give an example of the equation $\lfloor \frac{1}{|T|} \sum_{t \in T}^{} h_t(x_i) + \frac{1}{2} \rfloor$. When $70$ out of $|T|=100$ trees are voting for class 1, then we have $\lfloor\frac{1}{100}70 + 0.5\rfloor=\lfloor1.2\rfloor \rightarrow 1$. On the other side, if 25 out of $|T|=100$ trees are voting class 1 (the minority), then we have $\lfloor\frac{1}{100}25 + 0.5\rfloor=\lfloor0.75\rfloor \rightarrow 0$. Therefore, we are aiming to find the smallest $T'\subseteq T$, which will produce the same classification as the original $T$ trees.


\subsection{Reduction through Association Rules}

The first step of the reduction process begins by using association rules. Association rules~\cite{associationrules} mining is an unsupervised technique, which is used as a tool to extract knowledge from large datasets and explore relations between attributes. In association rules, the attributes are called items $I=\{i_{1},i_{2},\ldots ,i_{n}\}$. Each dataset contains sets of items, called itemsets $T=\{t_{1},t_{2},\ldots ,t_{m}\}$, where $t_i \subseteq I$. Using all possible items of a dataset, we can find all the rules $X\Rightarrow Y$, where $X,Y\subseteq I$. $X$ is called antecedent, while $Y$ is called consequent. In association rules the goal is to calculate the support and confidence of each rule in order to find useful relations. A simple observation is that $X$ is independent of $Y$ when the confidence is critically low. Furthermore, we can say that $X$ with high support, means it is probably very important.

\emph{But how can we use association rules in random forests?} We are going to do this at the path-level. The items $I$ will contain the features $F$ of our original dataset. The dataset $T$, which we are going to use to mine the association rules, will contain sets of features that represent each path $t_{i}=\{i_{j}|i_{j}=f_{j} | f_j \boxtimes v_k \in p_{i}, p_{i}\in P\}$. It is significant to mention that we keep only the presence of a feature in the path, and we discard its value $v_j$. Then, it is feasible to apply association rules techniques, like apriori algorithm~\cite{apriori}. 

The next step is to sort the association rules extracted by the apriori algorithm based on the ascending confidence score of each rule. For the rule $X \Rightarrow Y$, with the lowest confidence, we will take the $X$ and will add its items to the list of features. Afterwards, we are calculating the number of paths containing conjunctions, which are satisfied with the new feature set. If the amount of paths is at least half plus one paths of the total number of trees, we have found the reduced set of paths. \emph{We have a quorum!} Otherwise, we iterate and add more features from the next antecedent of the following rule. By using this technique, we reduce the number of features, and we have the new feature set $F' \subseteq F$. Reducing the features, can lead to a reduced set of paths too, because paths containing conjunctions with the redundant features will no longer be valid. Thus, for every path $p$ we have the following representation: \begin{equation}p=\{f_i \boxtimes v_j | f_i \in F', v_j \in \Re, \boxtimes \in \{\leq, >\}\}.\end{equation}

Illustrating this, for a toy dataset of four features $F = [f_1, f_2, f_3, f_4]$ and an RF model with five estimators $T = [t_1, t_2, t_3, t_4, t_5]$, for every instance $x$, from each $t_i \in T$ we can extract a path $p_i$. Supposing that for the instance $x$, we have five paths:

\begin{enumerate}[label=\subscript{p}{{\arabic*}}]
    \item if $f_1$ and $f_2$ and $f_4$ then Class\_A
    \item if $f_1$ and $f_3$ and $f_4$ then Class\_A
    \item if $f_1$ and $f_2$ and $f_4$ then Class\_A
    \item if $f_3$ and $f_4$ then Class\_A
    \item if $f_4$ then Class\_A
\end{enumerate}

Then, we can compute the association rules using apriori. Our objective is to create a set of features $F' \subset	 F$. We take the first rule $f_4 \Rightarrow (f_3, f_1$), the rule with the lowest confidence. This rule informs us that $f_4$, which has the highest support value, exists in $80\%$ of the paths without $(f_1, f_3)$. Thus, the first thing we add to our feature list is the antecedent of this rule, $f_4$. By adding the feature, we are counting how many paths can be satisfied with the features of $F' = [f_4]$. Only one path is valid ($p_5$), and is not enough because we need a quorum. Skipping all the association rules having the chosen features at their antecedents, the next rule we have is $f_1 \Rightarrow f_3$. $f_1$ has $0.6$ support value, and the rule has $0.33$ confidence. This means that in $66.6\%$ of paths containing $f_1$, the $f_3$ is absent. We add $f_1$ to the feature list and now the $F' = [f_1, f_4]$. With this feature list only the $p_5$ is activated again. Hence, we need another feature. The next rule we have is $f_3 \Rightarrow (f_1, f_4)$ with $0.4$ support of $f_3$ and confidence $0.2$. Adding $f_3$ now the paths $p_2, p_4$ and $p_5$ are valid.
    
In the aforementioned example, we achieved to reduce the features from four to three and the paths from five to three, as well. However, applying this method to datasets with plenty of features and models with more estimators, the reduction effect can be observed. Section~\ref{sec:exp} is seeking to explore that effect, through a set of experiments.

\subsection{Reduction through Clustering and Random Selection}
However, association rules may not be able to reduce the number of features and, consequently, the number of paths. In that case, a second reduction technique based on clustering is applied. Clustering is yet another group of unsupervised ML techniques, aside from the association rules. $k$-medoids~\cite{bauckhage2015numpy,kmedoidsori} is a well-known clustering algorithm, which considers as cluster's centre an existing element from the dataset. This element is called medoid. $k$-medoids, like other clustering techniques, needs a distance or a dissimilarity metric to find the optimum clusters. Thus, performing clustering to paths will require a path specific distance or dissimilarity metric.

We designed a path similarity metric in Algorithm~\ref{alg:similarityMetric}. This similarity metric is close to the distance metric introduced in iForest~\cite{iforest}, but eliminates some minor problems of the aforementioned. Parsing this algorithm, if a feature is absent from both paths, the similarity of these paths increases by 1. When a feature is present in both paths, the similarity increases by a value between 0 and 1, which is the intersection of the two ranges normalised by the union of the two ranges. The similarity metric is biased towards the absence of a feature from both paths because it always assigns one similarity point. However, this is a desirable feature for our goal of minimising the feature set.

\begin{algorithm}[h]
  \SetAlgoLined
\SetKwInOut{Input}{input}\SetKwInOut{Output}{return}
\Input{$p_i$, $p_j$, $feature\_names$, $min\_max\_feature\_values$}
\Output{$similarity_{ij}$}
 $s_{ij} \gets 0 $\\
 \For{$f$ in $feature\_names$}{
  \uIf{$f$ in $p_i$ \textbf{and} $f$ in $p_j$}{
   find $l_i$, $u_i$, $l_j$, $u_j$ lower and upper bounds\\
   $inter \gets min(u_i,u_j) - max(l_i,l_j)$\\
   $union \gets max(u_i,u_j) - min(l_i,l_j)$\\
   \If{$inter > 0$ \textbf{and} $union \neq 0$}{
     $s_{ij} \gets s_{ij} + inter/union$
   }
   }\ElseIf{$f$ not in $p_i$ \textbf{and} $f$ not in $p_j$}{
   $s_{ij} \gets s_{ij} + 1 $\\
  }
 }
 \KwRet{$s_{ij}/len(feature\_names) $}
 \caption{Path similarity metric}
 \label{alg:similarityMetric}
\end{algorithm}

In Algorithm~\ref{alg:kmeds}, we calculate the $k$-medoids and their clusters using the similarity metric of Algorithm~\ref{alg:similarityMetric}. Afterwards, we perform an ordering of the medoids based on the number of paths they cover in their clusters. Then, we collect paths from the larger clusters into a list, until we acquire at least a quorum. By summing larger clusters first, the possibility of feature reduction is increasing, because the paths inside a cluster tend to be more similar among them. Additionally, the biased similarity metric would cluster paths with less irrelevant features, leading to a subset of paths that are satisfied with a smaller set of features.

\begin{algorithm}[t]
\SetKwInOut{Input}{input}\SetKwInOut{Output}{return}
\Input{$similarity\_matrix$, $no\_of\_estimators$,\\
$paths$, $no\_of\_medoids$}
\Output{$paths$}
 $quorum \gets no\_of\_estimators/2 + 1$\\
 $m \gets kmedoids(similarity\_matrix, no\_of\_medoids)$\\
 $sorted\_m \gets sort\_by\_key(m, descending=True)$\\
 $count \gets 0$,
 $size \gets 0$,
 $reduced\_paths \gets []$\\
 \While{$size < quorum$ and $count < len(sorted\_m)$}{
     \For{$j$ in $m[sorted\_m[count]]$}{
        $reduced\_paths.append(paths[j])$
        }
     $count \gets count + 1$
     ,$size \gets len(reduced\_paths) $
 }
 \If{$size \geq quorum$}{
    $paths \gets reduced\_paths$\\
 }
 \KwRet{$paths$}
 \caption{Paths reduction through $k$-medoids clustering}
 \label{alg:kmeds}
\end{algorithm}

Performing clustering does not guarantee feature reduction, but there is a probability of an unanticipated reduction of the feature set. This procedure attempts to minimise the number of paths at least at the quorum. Unlike the association rules method, which may not accomplish to reduce features, clustering will significantly reduce the number of paths. By the end of the reduction process through clustering, random selection is applied to the paths to obtain the acceptable minimum number of paths, in case of reduction via clustering has not reached the quorum.

\subsection{Handling Categorical Features}
\label{sec:catf}
It is possible, even expected, to deal with a dataset containing categorical features. Of course, a transformation through OneHot or Ordinal~\cite{ordinal} encoding is used to make good use of these data. This transformed type of information will then be acceptable from ML systems. \emph{But is there any harm to interpretability caused by the use of encoding methods?} Sadly, yes! Using ordinal encoding will transform a feature like $country$$=$$[GR, UK,\dots]$ to $country$$=$$[0, 1, 2,\dots]$. As a result we lose the intelligibility of the feature. On the other hand, using OneHot encoding will increase dramatically the amount of features leading to over-length and incomprehensible interpretations by transforming the feature $country$ to $country\_GR$$=$$[0, 1]$, $country\_UK$$=$$[0, 1]$, and so forth. Since the encoding transformations are part of the feature engineering and are not invariable, we cannot construct a fully automated process to inverse transform the features into human interpretable forms within the interpretations.

Nevertheless, LionForests provides two automated encoding processes using either OneHot or Ordinal encoding and their inverse transformation for the interpretation extraction. Feature-ranges of Ordinal encoded data transform like $(1$$\leq$$country$$\leq$$2)$$\to$$(country$$=$$[UK, FR])$, while feature-ranges of OneHot encoded data $(0.5$$\leq$$country\_UK$$\leq$$1)$$\to$$(country$$=$$UK)$ and feature-ranges like $(0$$\leq$$country\_GR$$\leq$$0.49)$ are removed. The excluded OneHot encoded features for the categorical feature should appear to the user as possible alternative values. If a feature reduction method reduces one of the encoded OneHot features, it will not appear in the feature's list of alternative values, but in the list of values that do not influence the prediction. For this reason, the categorical features will appear in the interpretations with a notation `c' like `$categorical\_feature^c = value$'. Depending on the application and the interpretation, the user will be able to request a list of alternative values or will be able to simply hover over the feature to reveal the list. Section~\ref{sbs:adult} is showcasing transformations of OneHot encoded features, as well as one example of a OneHot encoded feature's alternative values list. 

\subsection{Interpretation Composition}
These processes are part of the LionForests technique, which, in the end, produces an interpretation in the form of a feature-range rule. Lastly, LionForests combines the ranges of the features in the reduced feature set to a single natural language rule. The order of appearance of the feature ranges in the rule is determined by using a global interpretation method, such as the SHAP TreeExplainer~\cite{shapTrees} or the Scikit~\cite{sklearn} RF model's built-in \emph{feature importance} attribute, for smaller and larger datasets, respectively. One notable example of an interpretation is the following:
\begin{equation}
\centering
    \text{`if } 0 \leq f_1 \leq 0.5 \text{ and } -0.5 \leq f_3 \leq 0.15 \text{ then class\_} A\text{'.}
\end{equation}
We interpret this feature-range rule like that: ``As long as the value of the $f_1$ is between the ranges 0 and 0.5, and the value of $f_3$ is between the ranges -0.5 and 0.15, the system will classify this instance to class A. If the value of $f_1$, $f_3$ or both, surpass the limits of their ranges then the prediction may change. Note that the features are ranked through their influence''. This type of interpretation is comprehensible and human-readable. Thus, if we manage to keep them shorter, then they could be an ideal way to explain an RF model. A way to encounter an over-length rule could be to hide the last $n$ feature-ranges, which they will be the least important due to the ordering process. At the same time, users will have the ability to expand their rules to explore the feature-ranges. However, we do not completely exclude those features, but we are only hiding them because otherwise, we would affect the correctness of both the explanation and the prediction. An example is showcased in Section~\ref{sbs:adult}.

\renewcommand{\thetable}{2}
\begin{table*}[b]
\centering
\resizebox{\textwidth}{!}{%
\begin{tabular}{ccc|c|c|c|c|c|c|cc}
\hline
\multicolumn{3}{|c|}{\textbf{Reduction Technique}} & \multicolumn{6}{c|}{\textbf{Reduction \%}} & \multicolumn{2}{c|}{\textbf{Mean Reduction \%}} \\ \hline
\multicolumn{1}{|c|}{Association Rules} & \multicolumn{1}{c|}{Clustering} & Random Based & Feature & Path & Feature & Path & Feature & Path & \multicolumn{1}{c|}{Feature} & \multicolumn{1}{c|}{Path} \\ \hline
\multicolumn{1}{|c|}{\checkmark} & \multicolumn{1}{c|}{\checkmark} & \checkmark & \textbf{30.85}\smaller{$\pm5.1e^{-2}$} & \textbf{49.47}\smaller{$\pm5.5e^{-15}$} & \textbf{21.37}\smaller{$\pm0.0$} & \textbf{43.41}\smaller{$\pm0.0$} & \textbf{10.03}\smaller{$\pm1.5e^{-2}$}  & \textbf{44.18}\smaller{$\pm5.5e^{-15}$} & \multicolumn{1}{c|}{\textbf{20.75}} & \multicolumn{1}{c|}{\textbf{45.69}} \\ \hline
\multicolumn{1}{|c|}{-} & \multicolumn{1}{c|}{\checkmark} & \checkmark & 2.15\smaller{$\pm1.8e^{-1}$} & \textbf{49.47}\smaller{$\pm5.5e^{-15}$} & $8.6e^{-3}$\smaller{$\pm1.3e^{-2}$} & \textbf{43.41}\smaller{$\pm0.0$} & \textbf{10.03}\smaller{$\pm1.5e^{-2}$} & \textbf{44.18}\smaller{$\pm5.5e^{-15}$} & \multicolumn{1}{c|}{4.06} & \multicolumn{1}{c|}{\textbf{45.69}} \\ \hline
\multicolumn{1}{|c|}{\checkmark} & \multicolumn{1}{c|}{-} & \checkmark & 30.70\smaller{$\pm0.0$} & \textbf{49.47}\smaller{$\pm5.5e^{-15}$} & \textbf{21.37}\smaller{$\pm0.0$} & \textbf{43.41}\smaller{$\pm0.0$} & 9.11\smaller{$\pm1.4e^{-2}$} & \textbf{44.18}\smaller{$\pm5.5e^{-15}$} & \multicolumn{1}{c|}{20.39} & \multicolumn{1}{c|}{\textbf{45.69}} \\ \hline
\multicolumn{1}{|c|}{\checkmark} & \multicolumn{1}{c|}{\checkmark} & - & 30.84\smaller{$\pm4.4e^{-2}$} & 48.56\smaller{$\pm2.4e^{-2}$} & \textbf{21.37}\smaller{$\pm0.0$} & 42.88\smaller{$\pm2.2e^{-2}$} & 7.82\smaller{$\pm1.4e^{-2}$} & 36.26\smaller{$\pm2.6e^{-2}$} & \multicolumn{1}{c|}{20.01} & \multicolumn{1}{c|}{42.57} \\ \hline
\multicolumn{1}{|c|}{\checkmark} & \multicolumn{1}{c|}{-} & - & 30.70\smaller{$\pm0.0$} & 27.02\smaller{$\pm0.0$} & \textbf{21.37}\smaller{$\pm0.0$} & 32.67\smaller{$\pm0.0$} & 0.0\smaller{$\pm0.0$} & 0.0\smaller{$\pm0.0$} & \multicolumn{1}{c|}{17.36} & \multicolumn{1}{c|}{19.90} \\ \hline
\multicolumn{1}{|c|}{-} & \multicolumn{1}{c|}{\checkmark} & - & 2.12\smaller{$\pm1.3e^{-1}$} & 47.57\smaller{$\pm3.6e^{-2}$} & $5.7e^{-3}$\smaller{$\pm1.1e^{-2}$} & 41.98\smaller{$\pm7.4e^{-2}$} & 7.82\smaller{$\pm1.4e^{-2}$}  & 36.26\smaller{$\pm2.6e^{-2}$} & \multicolumn{1}{c|}{3.32} & \multicolumn{1}{c|}{41.94} \\ \hline
\multicolumn{1}{|c|}{-} & \multicolumn{1}{c|}{-} & \checkmark & 0.0\smaller{$\pm0.0$} & \textbf{49.47}\smaller{$\pm5.5e^{-15}$} & $2.9e^{-3}$\smaller{$\pm8.6e^{-3}$} & \textbf{43.41}\smaller{$\pm0.0$} & 9.11\smaller{$\pm1.4e^{-2}$} & \textbf{44.18}\smaller{$\pm5.5e^{-15}$} & \multicolumn{1}{c|}{3.04} & \multicolumn{1}{c|}{\textbf{45.69}} \\ \hline
 &  &  & \multicolumn{2}{c|}{Banknote Authentication} & \multicolumn{2}{c|}{Heart Disease} & \multicolumn{2}{c|}{Adult Census} &  &  \\ \cline{4-9}
\end{tabular}%
}
\caption{Feature and path reduction ratios on the three datasets. \checkmark(\#) the order of the applied techniques}
\label{tab:featurePathRatios}
\end{table*}

\section{EXPERIMENTAL RESULTS}
\label{sec:exp}

We first discuss the setup of our experiments. Then, we present quantitative results, followed by qualitative results for the explanation of particular instances.

\subsection{Experimental Setup}

The implementation of LionForests as well as of all the experiments in this paper is available in the LionLearn repository at GitHub\footnote{\href{https://github.com/intelligence-csd-auth-gr/LionLearn}{https://github.com/intelligence-csd-auth-gr/LionLearn}}.

Our experiments were conducted on the following three tabular binary classification datasets: Banknote authentication~\cite{ucidata}, Heart (Statlog)~\cite{ucidata} and Adult Census~\cite{adultDataset}. The top part of Table~\ref{tab:datasets} shows the number of instances and features of each dataset. Particularly, Adult Census contains 6 numerical and 8 categorical features. The eight categorical features are transformed through OneHot encoding into 74 numerical, resulting in a total of 80 features.

We used the RandomForestClassifier from Scikit-learn~\cite{sklearn} and the MinMaxScaler with feature range $[-1,1]$. In order to work with optimised models, a 10-fold cross-validation grid search was carried out on each dataset, using the following set of parameters and values: max\_depth \{1, 5, 7, 10\}, max\_features \{`sqrt', `log2', 75\%, None\footnote{None = all features}\}, min\_samples\_leaf \{1, 2, 5, 10, 10\%\}, bootstrap \{True, False\}, n\_estimators \{10, 100, 500, 1000\}. The parameters' values that achieved the best F$_1$ score, along with the score itself are shown in the middle and bottom part of Table~\ref{tab:datasets}, respectively. 

\renewcommand{\thetable}{1}
\begin{table}
    \centering
    \resizebox{0.4\textwidth}{!}{%
    \begin{tabular}{rccc}
                &  Banknote & Heart (Statlog) & Adult Census \smallskip\\
         \hline \noalign{\smallskip}
     instances  & 1372  & 270   & 48842 \\
     features   & 4     & 13    & 14 (80)    \smallskip\\
         \hline \noalign{\smallskip}
    max depth   & 10    & 5     & 10 \\
    max features& 0.75  & `sqrt'& `sqrt' \\
min sample leaf & 1     & 5     & 1 \\
    bootstrap   & True  & False & False \\
    estimators  & 500   & 500   & 100 \smallskip\\
    \hline \noalign{\smallskip}
     $F_1 \%$   & 99.43 & 81.89 & 88.71 \\
    \end{tabular}
    }
    \caption{For each of the three datasets: main statistics (top), optimal parameter values of the Random Forest (middle), corresponding $F_1$ score (bottom)}
    \label{tab:datasets}
\end{table}

\subsection{Quantitative Results}

For each dataset, we train one RF model, using all data, with the parameters shown in Table~\ref{tab:datasets}. Then we apply LionForests to all instances of each dataset and report the mean feature and path reduction in Table~\ref{tab:featurePathRatios}.

In terms of path reduction, we notice that among the three reduction techniques, random selection consistently leads to the best results across all datasets, achieving an average reduction of 45.69\%. Clustering is the second-best method with only slightly worse results in the first two datasets, but much worse results in the Adult Census dataset, achieving an average of 41.94\% reduction. Association rules is the worst technique, achieving zero reduction in the Adult Census dataset and an average of 19.90\% reduction. Combining random selection with other techniques does not lead to improved results.

With respect to feature reduction, we find that the association rules strategy leads to the best results in the first two datasets, reaching an average reduction of 26.04\%. The other two techniques achieve negligible reduction in these two datasets. In banknote authentication combining the three methods improves the reduction slightly from 30.70\% to 30.85\%. In Adult Census on the other hand, association rules achieve zero reduction of features similarly to paths. The best result in this dataset is achieved by random selection, followed closely by clustering. Combining these two techniques improves slightly the reduction to 10\%.

The weak performance of association rules in Adult Census is related to the small number of estimators (100) and the huge number of features (80). In particular, each of the estimators can have a maximum of eight features, as resulted from the grid search (max features = `sqrt'). Thus, the overlapping features among the different paths are fewer, and by extension, techniques relying on feature selection, like association rules, cannot perform the feature reduction, as well as path reduction. However, LionForests is an ensemble of techniques, and we have revealed with these quantitative experiments the importance of each part. We may infer that the LionForests strategy is considerably effective in reducing both features and paths. 

\subsection{Qualitative Results}
LionForests technique provides consistent and robust rules which are more indisputable from other interpretations because they are more compact, have broader ranges for the features, while at the same time present categorical data in a human-comprehensible form. In this section, we provide an example from each dataset to demonstrate how RF models can be interpreted efficiently.

\subsubsection{Banknote Authentication}
We observe that the complete methodology, incorporating association rules, clustering and random selection reduction, achieves the highest performance on both feature and path reduction for the first dataset, Banknote Authentication. We also provide an example of a pair of explanations (1) without and (2) with LionForests:

\begin{enumerate} 
    \item `if $2.4 \leq variance \leq 6.83$ and $-3.13\leq skew \leq-2.30$ and $1.82\leq curtosis \leq2.13$ and $-0.64\leq entropy \leq0.73$ then fake banknote'
    \item `if $2.4\leq variance \leq6.83$ and $-1.60\leq curtosis \leq17.93$ then fake banknote'
\end{enumerate}

The reduced rule (2) is smaller than the original by two features. In addition, the ``curtosis'' feature has a wider range. The instance has a value of $1.92$ for the ``curtosis'' feature, and in the original rule this value is marginal for the very narrow range $1.82\leq curtosis \leq2.13$ indicating that a small change may lead to a different outcome, but this is not the case for the reduced rule as well. In addition, changing the skew value from $-2.64$ to $-4$, which is outside the range of the feature in the original rule, will not change the prediction and will produce the same reduced range rule. We observe the same result when we change the value of the ``entropy'' feature, as well as when we tweak both ``skew'' and ``entropy''.

\subsubsection{Heart Disease}
Again, with LionForests we achieve both higher feature and path reduction ratios. There are thirteen features included in this particular dataset and, as a result, the interpretations may have a total of thirteen features. In fact, in this case the reduction of features is essential to provide comprehensible interpretations. We choose an example, and we present the original rule (1) and the reduced rule (2):

\begin{enumerate}
    \item `if $6.5 \leq reversable\ defect \leq 7.0$ and $3.5 \leq chest\ pain \leq 4.0$ and $0.0 \leq number\ of\ major\ vessels \leq 0.5$ and $1.55 \leq oldpeak \leq 1.7$ and $0.5 \leq exercise\ induced\ angina \leq 1.0$ and $128.005\leq maximum\ heart\ rate\ achieved \leq 130.998$ and $1.5 \leq the\ slope\ of\ the\ peak\ exercise \leq 2.5$ and $sex^c = Male$ and $184.999 \leq serum\ cholestoral \leq 199.496$ and $29.002 \leq age \leq 41.497$ and $0.0 \leq resting\ electrocardiographic\ results \leq 0.5$ and $119.0 \leq resting\ blood\ pressure \leq 121.491$ and $0.0 \leq fasting\ blood\ sugar \leq 0.5$ then presence'
    \item `if $6.5 \leq reversable\ defect \leq 7.0$ and $3.5 \leq chest\ pain \leq 4.0$ and $0.0 \leq number\ of\ major\ vessels \leq 0.5$ and $1.55 \leq oldpeak \leq 1.7$ and $0.5 \leq exercise\ induced\ angina \leq 1.0$ and $128.005 \leq maximum\ heart\ rate\ achieved \leq 133.494$ and $1.5 \leq the\ slope\ of\ the\ peak\ exercise \leq 2.5$ and $184.999 \leq serum\ cholestoral \leq 199.496$ and $119.0 \leq resting\ blood\ pressure \leq 121.491$ then presence'
\end{enumerate}

The reduced rule is shorter than the original rule by four features. In addition, the ``maximum heart rate achieved'' feature has a wider range in the reduced rule. Changing the ``sex'' value from `Male' (1) to `Female' (0) did not change the reduced rule at all. We tweak ``age'' value from 35 to 15 and again the reduced rule remains the same. Thus, features like ``age'', ``sex'', ``resting electrocardiographic results'' and ``fast blood sugar'', cannot influence the prediction.

\subsubsection{Adult Census} 
\label{sbs:adult}

Finally, we present a pair of explanations provided by LionForests for an instance of Adult Census, without (1) and with (2) reduction:

\begin{enumerate}
    \item `if $marital$ $status^c = Married$ and $sex^c = Female$ and $education^c = HS$ $grad$ and $workclass^c = Private$ and $94721 \leq$ $fnlwgt$ $\leq 161182$ and \textbf{$47 \leq$ $age$ $\leq 53$} and $15 \leq$ $hours$ $per$ $week$ $\leq 25$ and $native$ $country^c = Jamaica$ and \emph{[other 2 feature-ranges]} then income >50K'
    \item `if $marital$ $status^c = Married$ and $sex^c = Female$ and $education^c = HS\_grad$ and $workclass^c = Private$ and $87337 \leq fnlwgt \leq 382719$ and \textbf{$47 \leq age \leq 63$} and $15 \leq hours$ $per$ $week \leq 99$ and $native$ $country^c = Jamaica$ and \emph{[other 2 feature-ranges]} then income >50K'
\end{enumerate}

The reduced rule is thirteen features smaller than the original rule. This is not directly obvious because some OneHot categorical features are not presented. For example, only valid features such as ``marital\_status\_Married'' and ``education\_HS\_Grad'' are presented as described in Section~\ref{sec:catf}. Furthermore, we observe that the ranges of ``age'', ``fnlwft'' and ``hours per week'' are broader. Specifically, ``age'' range from $[47,53]$ increased to $[47,63]$, while ``hours per week'' range from $[15,25]$ expanded to $[15,99]$. Moreover, we can explore the categorical feature's ``native country'' alternative values. In Table~\ref{tab:nativevalues}, the first list refers to the values that may change the prediction of an instance, while the second shows the values that cannot affect the prediction. In the original rule this type of information was not available because the non-affecting values were present.

\renewcommand{\thetable}{3}
\begin{table}
\centering
\resizebox{0.49\textwidth}{!}{%
\begin{tabular}{|c|c|}
\hline
\multicolumn{2}{|c|}{\textbf{Possible values of ``native\_country'', which they}} \\
\textbf{may affect the prediction} & \textbf{preserve the prediction} \\ \hline
\begin{tabular}[c]{@{}c@{}}`Mexico', `United-States', `Canada', \\ `Philippines', `England', `Thailand', \\`Japan', `China', `Dominican-Republic', \\ `Germany', `South', `Columbia', `Italy',\\`Puerto-Rico', `Vietnam', `Cambodia', \\ `Ireland', `Taiwan', `Portugal', `Laos',\\`Yugoslavia', `Nicaragua', `Scotland'\\\end{tabular} & \begin{tabular}[c]{@{}c@{}}`India', `France',\\  `El-Salvador', `Iran', \\`Cuba', `Haiti', \\`Guatemala', `Peru', \\ `Trinidad\&Tobago', \\ `Honduras' \end{tabular} \\ \hline
\end{tabular}
}
\caption{List of values affecting or not the classification of an instance}
\label{tab:nativevalues}
\end{table}

\section{CONCLUSION}
Providing helpful explanations is a challenging task. Providing comprehensible and undeniable explanations is even tougher. Since model-agnostic approaches produce approximations, i.e. the nearest optimal ones but not the optimal explanations, the attempt to indifferently interpret each black-box model will not lead to the desired outcome. In this work, we introduced a model-specific local-based approach for obtaining real interpretations of random forests predictions. Other works~\cite{moore,iforest} attempt to provide explanations of this form, but they do not try to make them more comprehensible, either indisputable. A user may not be familiar with iForest's~\cite{iforest} visualisation tool. Besides, an interpretation containing a lot of features with narrow ranges~\cite{moore} may lead to incomprehensible and untrustworthy rules. The proposed technique, LionForests, will provide users with small rules as interpretations in natural language, which by widening the feature ranges will be more reliable and trustworthy as well. We use classic unsupervised methods such as association rules and $k$-medoids clustering to achieve feature and path reduction.

Nevertheless, LionForests is not a complete solution either, since it is not appropriate for tasks of model inspection. For example, if a researcher is working to build a reliable and stable model aiming for the highest performance, a visualisation tool like iForest may be preferred. This approach is the best and easiest way of providing an interpretation to a non-expert user. Lastly, by reducing the number of paths to the quorum to minimise the features and at the same time to increase the features' ranges, the outcome would be a discounted probability of the classification of the instance, which poses questions about the prediction's reliability. This can be counter-attacked by introducing a threshold parameter to the reduction effect, requiring the algorithm to retain at least a specific percentage of the paths.

Future research will explore the impact of tuning parameters, such as the number of estimators or the maximum number of features, on the reduction of features and paths. We also aim to apply LionForests to different tree ensembles, rather than random forests, as well as to various datasets and data types. FP\-Growth~\cite{fpgrowth}, and its variant FP\-Max~\cite{fpmax}, will be tested against the Apriori algorithm. In addition, we will consider the possibility of adapting LionForests to other tasks, such as multi-class or multi-label classification, as well as regression. We will also explore the possibility of using LionForests interpretations to provide descriptive narratives through counter-examples. Ultimately, by means of a qualitative, human-oriented analysis, we will try to explore this promising method in order to prove its intelligibility and its necessity as a foundation for human-centred artificial intelligence systems based on interpretable ML methods.

\ack This paper is supported by the European Union's Horizon 2020 research and innovation programme under grant agreement No 825619, AI4EU Project\footnote{\url{https://www.ai4eu.eu}}.

\end{document}